\documentclass[10pt, a4paper]{article}

\usepackage{graphicx}
\usepackage{latexsym}

\usepackage{booktabs}
\usepackage{multirow}
\usepackage{makecell}
\usepackage{amsmath}
\usepackage{amsfonts,bm}
\usepackage{tikz}
\usepackage{lipsum}

\usetikzlibrary{shapes.geometric}

\DeclareMathOperator*{\softmax}{softmax}

\usepackage[]{lrec-coling2024} 

\title{A Federated Learning Approach to Privacy Preserving \\ Offensive Language Identification}

\name{Marcos Zampieri\textsuperscript{1}, Damith Premasiri\textsuperscript{2}, Tharindu Ranasinghe\textsuperscript{3}, } 

\address{\textsuperscript{1}George Mason University, USA, 
         \textsuperscript{2}Lancaster University, UK,
         \textsuperscript{3}Aston University, UK \\
          \texttt{mzampier@gmu.edu}\\ }

\abstract{
The spread of various forms of offensive speech online is an important concern in social media. While platforms have been investing heavily in ways of coping with this problem, the question of privacy remains largely unaddressed. Models trained to detect offensive language on social media are trained and/or fine-tuned using large amounts of data often stored in centralized servers. Since most social media data originates from end users, we propose a privacy preserving decentralized architecture for identifying offensive language online by introducing Federated Learning (FL) in the context of offensive language identification. FL is a decentralized architecture that allows multiple models to be trained locally without the need for data sharing hence preserving users' privacy. We propose a model fusion approach to perform FL. We trained multiple deep learning models on four publicly available English benchmark datasets (AHSD, HASOC, HateXplain, OLID) and evaluated their performance in detail. We also present initial cross-lingual experiments in English and Spanish. We show that the proposed model fusion approach outperforms baselines in all the datasets while preserving privacy.
 \\ \newline \Keywords{federated learning, offensive language identification, privacy} }

\begin{document}

\maketitleabstract

\pagestyle{empty}

\section{Introduction}

NLP systems relying on modern deep learning paradigms are trained on very large amounts of data. In several applications and domains (e.g., social media), most data used to train machine learning models comes from end users. Such confidential data often cannot be shared without compromising users' privacy. This is an important concern for organizations that handle large amounts of confidential data, such as financial institutions, healthcare facilities, law firms, and many others. With the widespread use of personal computing devices (e.g., PCs, smartphones, and virtual assistants), data privacy also became a great concern to individuals, which motivated several countries to pass legislation aiming to protect users' privacy such as the European Union General Data Protection Regulation (GDPR)\footnote{\url{https://gdpr.eu/}} and the Swiss Datenschutzgesetz (DSG).\footnote{\url{https://www.edoeb.admin.ch/edoeb/de/home/datenschutz/ueberblick/datenschutz.html}}


The need for privacy-preserving machine learning models that can handle confidential data while protecting organizations' and users' privacy emerges from this situation. To address this important challenge, Federated Learning (FL) has become an increasingly popular machine learning paradigm \cite{mcmahan2017communication} as it allows us to train robust machine learning models across multiple devices or servers without exchanging data. In FL, multiple clients work together under the coordination of a central server. Each client’s data is stored locally and not exchanged among clients or with the central server. FL, therefore, offers the possibility of training robust machine learning models on large numbers of decentralized local data repositories without compromising privacy. FL models have been successfully applied in a wide range of applications in computer networks \cite{lim2020federated}, computer vision \cite{yan2021experiments}, information retrieval \cite{wang2021efficient}, NLP \cite{chen2019federated}, and many others. 

\begin{table*}
\centering
\setlength{\tabcolsep}{5pt}
\scalebox{0.9}{
\begin{tabular}{l|cc|cc|l}
\hline
& \multicolumn{2}{c|}{\bf Training} & \multicolumn{2}{c|}{\bf Testing} &  \\
\bf Dataset  & \bf Inst. & \bf OFF \% & \bf Inst.  & \bf OFF \% & \bf Data Sources \\ 
\hline
AHSD \cite{davidson2017}  & 19,822 & 0.83 & 4,956 & 0.82 & Twitter \\ 
HASOC \cite{mandl2020} & 5,604 & 0.36 & 1,401  & 0.35 & Twitter, Facebook    \\ 
HateXplain  \cite{mathew2020hatexplain}  & 11,535 & 0.59  & 3,844  & 0.58 & Twitter, Gab  \\ 
OLID \cite{OLID} & 13,240 & 0.33  & 860 & 0.27 & Twitter   \\
\hline
\end{tabular}
}
\caption{The four datasets, including the number of instances (Inst.) in the training and testing sets, the OFF \% in each set and the data source.}
\label{tab:data}
\end{table*}

In this paper, we explore the use of FL in offensive language identification online through a model fusion technique \cite{choshen2022fusing}. Datasets containing the various forms of offensive speech (e.g., hate speech, cyberbullying, etc.) are sensitive in nature, which creates an interesting use case for FL. The use of FL and other privacy-preserving paradigms allows social media platforms to work together to solve this important issue without the need to exchange confidential information, thus preserving users' privacy. While FL has recently started to be explored in NLP \cite{chen2019federated,lin2022fednlp}, including the workshop on Federated Learning for NLP (FL4NLP) at ACL-2022 \cite{fl4nlp-2022-federated}, to the best of our knowledge, no studies have yet explored the use of FL in the context of offensive language identification. Our work fills this gap by introducing FL in the context of offensive language identification online and by providing the community with an evaluation of FL methods using four publicly available English offensive language benchmark datasets presented in Section \ref{sec:data}.

One recent study \cite{gala2023federated} proposed FL in offensive language identification, but it lacks the consideration of combining different data. Their architecture solely focuses on distributed training on the same dataset with multiple clients and evaluating \textit{fedopt \cite{reddi2021adaptive}, fedprox \cite{anonymous2019federated}} algorithms to optimise the global model. Our main focus in this study is on combining multiple models using FL, which could identify offensive content in different data. 




\section{Related Work}

\noindent \textbf{\textit{Offensive Language Identification}} The task of automatically identifying offensive language online has been substantially explored in the literature \cite{macavaney2019hate,melton2020hate,zia2022improving,weerasooriya2023vicarious}. Multiple types of offensive content have been addressed, such as {\em aggression}, {\em cyberbulling}, and {\em hate speech} using classical machine learning classifiers (e.g., Support Vector Machines) \cite{malmasi2017detecting,malmasi2018challenges}, neural networks \cite{gamback2017using,djuric2015hate,hettiarachchi-ranasinghe-2019-emoji}, pre-trained general-purpose transformer-based language models \cite{ranasinghe-etal-2020-multilingual,ranasinghemudes}, and fine-tuned language models on offensive language datasets \cite{caselli2020hatebert,sarkar2021fbert}. The vast majority of studies addressed offensive content in English and other widely-spoken resource-rich languages such as Arabic \cite{mubarak2020arabic}, Portuguese \cite{fortuna2019hierarchically} and Turkish \cite{coltekin2020} while a few studies dealt with low-resource languages \cite{fiser2017,gaikwad2021cross,raihan2023offensive}. Multiple competitions on this topic have been organized creating important benchmark datasets such as OffensEval \cite{offenseval,zampieri-etal-2020-semeval}, HASOC \cite{mandl2020,modha2021overview,satapara2022overview}, TRAC \cite{kumar2018benchmarking,kumar-etal-2020-evaluating}, and HatEval \cite{basile2019semeval}. While substantial progress has been made in the past few years, to the best of our knowledge, none of the aforementioned studies or competitions has addressed the question of data privacy.  

\vspace{2mm}

\noindent \textbf{\textit{Federated Learning in NLP}} With the goal of preserving users' data privacy, FL architectures have been extensively studied in a variety of domains \cite{wang2021efficient} in the past several years. Only more recently, however, FL has been explored for text and speech processing \cite{lin2022fednlp,silva2023fedperc,zhang2023fedlegal,che2023federated}. Recent workshops co-located with top-tier conferences confirm this growing interest in FL and privacy in general. The workshop on Privacy in Natural Language Processing (PrivateNLP) \cite{privatenlp-2022-privacy}, which is currently in its fourth edition, addressed the interplay between NLP and data privacy while the aforementioned FL4NLP workshop \cite{fl4nlp-2022-federated} co-located with ACL-2022 was the first workshop organized focusing exclusively on FL for NLP. Most papers presented in the workshop, however, dealt with language modelling and learning representation rather than with downstream tasks and applications such as offensive language identification. As we mentioned before, a recent study applied different FL strategies in offensive language identification \cite{gala2023federated}. However, their study focuses on distributed training on the same dataset \cite{anonymous2019federated}.

\section{Data}
\label{sec:data}

We use four popular publicly available datasets containing English data summarized in Table \ref{tab:data}. As the datasets were annotated using different guidelines and labels, following the methodology described in previous work \cite{ranasinghe-etal-2020-multilingual}, we map all labels to OLID level A \cite{OLID}, which contains the labels offensive (OFF) vs. not offensive (NOT). We choose OLID due to the flexibility provided by its general three-level hierarchical taxonomy below, where the OFF class contains all types of offensive content, from general profanity to hate speech, while the NOT class contains non-offensive examples. The OLID taxonomy is presented next:

\begin{figure*}[!ht]
\centering
\includegraphics[scale=0.52]{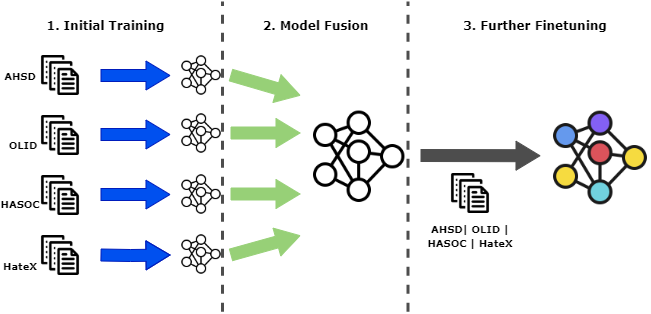}
\caption{The three stages of the FL pipeline in the proposed fused model.}
\label{fig:federated_learning}
\end{figure*}

\begin{itemize}
    \item {\bf Level A:} Offensive (OFF) vs. Non-offensive (NOT).
    \item {\bf Level B:} Classification of the type of offensive (OFF) tweet - Targeted (TIN) vs. Untargeted (UNT).
    \item {\bf Level C:} Classification of the target of a targeted (TIN) tweet - Individual (IND) vs. Group (GRP) vs. Other (OTH).
\end{itemize}

\noindent In the OLID taxonomy, offensive (OFF) posts targeted (TIN) at an individual are often cyberbulling whereas offensive (OFF) posts targeted (TIN) at a group is often hate speech. 

\vspace{2mm}

\noindent \textbf{AHSD} \cite{davidson2017} is one of the most popular hate speech datasets available. The dataset contains data retrieved from Twitter and it was annotated using crowdsourcing. The annotation taxonomy contains three classes; Offensive, Hate, and Neither. We conflate Offensive and Hate under a class OFF while neither class corresponds to OLID's NOT class.

\vspace{2mm}

\noindent \textbf{OLID} \cite{OLID} is the official dataset of the SemEval-2019 Task 6 (OffensEval) \cite{offenseval}. It contains data from Twitter annotated with a three-level hierarchical annotation in which level A classifies posts into offensive and not offensive; level B differentiates between targeted pots (insults and threats) and untargeted posts (general profanity); and level C classifies them into three targets: individual, group, or other. We adopt the labels in OLID level A as our classification labels.

\vspace{2mm}

\noindent \textbf{HASOC} \cite{mandl2020} is the dataset used in the HASOC shared task 2020. It contains posts retrieved from Twitter and Facebook. The upper level of the annotation taxonomy used in HASOC is the same as OLID's level A, which allows us to directly use the same labels in our models. 

\vspace{2mm}

\noindent \textbf{HateXplain} \cite{mathew2020hatexplain} is a recent dataset collected for the explainability of hate speech. It contains both token- and post-level annotation of Twitter and Gab posts. The annotation taxonomy contains three classes; hate speech, offensive speech, and normal. Following the annotation guidelines of OLID \cite{OLID}, we mapped the hate speech and offensive speech classes to offensive (OFF) and normal class to not offensive (NOT).

\section{Methodology}

The proposed FL pipeline contains three steps depicted in Figure \ref{fig:federated_learning}. We describe these steps below.

\begin{figure}[!ht]
\centering
\includegraphics[scale=0.44]{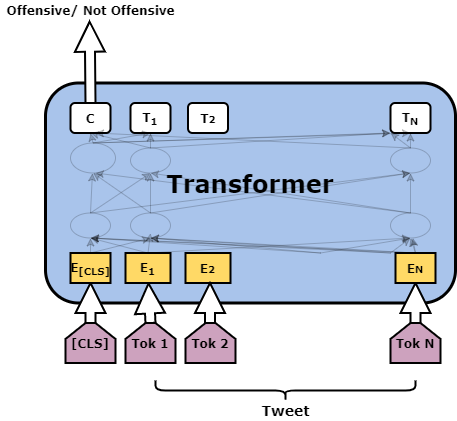}
\caption{A sample transformer model for offensive language identification \cite{ranasinghe-etal-2020-multilingual} predicting offensive and not offensive labels.}
\label{fig:transformer}
\end{figure}

\noindent \textbf{\textit{Initial Model Training}}
Transformer models have achieved state-of-the-art performance in many NLP tasks \cite{devlin2019bert}, including offensive language identification \cite{ranasinghe2019brums,sarkar2021fbert}. Therefore, our methodology in this paper builds around pre-trained transformers. For the text classification tasks such as offensive language identification, we use the pre-trained transformer models by utilizing the hidden representation of the classification token (\textsc{CLS}) as shown in Figure \ref{fig:transformer}. For this task, we implemented a softmax layer on top of the \textsc{CLS} token, i.e., the predicted probabilities are $\bm y^{(B)}=\softmax(W\bm h + b)$, where $W\in\mathbb{R}^{k\times d}$ is the softmax weight matrix, and $k$ is the number of labels. which in our case is always equal to two. 

\newpage

We used this text classification architecture to build separate models for each dataset that we introduced in the previous section. We trained the model using the training sets of each dataset. We employed a batch-size of 16, Adam optimiser with learning rate $4\mathrm{e}{-5}$, and a linear learning rate warm-up over 10\% of the training data. During the training process, the parameters of the transformer model and the parameters of the subsequent layers were updated. The models were evaluated while training using an evaluation set that had one-fifth of the rows in training data. We performed early stopping if the evaluation loss did not improve over three evaluation steps. All the models were trained for three epochs. 

\begin{table*}[!ht]
    \centering
    \scalebox{0.92}{
        \begin{tabular}{c | c | c c c c | l  }
        \hline
             \bf Dataset & \bf Approach & & \multicolumn{2}{c}{\bf Models} & & \bf Macro F1 \\

            \hline
            \multirow{4}{*}{AHSD} & non-fused & AHSD &  & - & - & 0.931 \textpm 0.01\\
            & fusion with FT & AHSD & OLID & - & - & 0.921 \textpm 0.00\\
            & fusion without FT & AHSD & OLID & - & - & 0.866 \textpm 0.00\\
            & ensemble & AHSD & OLID & - & - & 0.845 \textpm 0.01\\
            \hline
                 \multirow{4}{*}{OLID} & non-fused & - & OLID & - & - & 0.854 \textpm 0.00\\
            & fusion with FT & AHSD & OLID & - & - & 0.837 \textpm 0.03\\
            & fusion without FT & AHSD & OLID & - & - & 0.836 \textpm 0.00\\
            & ensemble &  & OLID & - & HateX & 0.785 \textpm 0.04\\

            \hline
                   \multirow{4}{*}{HASOC} & non-fused & - &  - & HASOC & - & 0.798 \textpm 0.01\\
                 & fusion without FT & AHSD & OLID & HASOC & - & 0.770 \textpm 0.01 \\
            & fusion with FT & AHSD & OLID & HASOC & - & 0.754 \textpm 0.07\\
       
            & ensemble & AHSD &  & HASOC & - & 0.647\textpm 0.02\\
            \hline
                   \multirow{4}{*}{HateX} & non-fused & - &  & - & HateX & 0.795 \textpm 0.01\\
            & fusion with FT & AHSD & - & - & HateX & 0.777 \textpm 0.00\\
            & fusion without FT & AHSD & OLID & - & HateX & 0.772 \textpm 0.01\\
            & ensemble & - & - & HASOC & HateX & 0.654 \textpm 0.01\\
            \hline

        \end{tabular}
    }
    \caption{The best result for each dataset for each approach; non-fused models, fused models with fine-tuning (FT), fused models without finetuning and ensemble. We only report the results with fBERT. The results are ordered from Macro F1.}
    \label{tab:best_results}
\end{table*}

We repeated this process with two popular pre-trained transformer models; \textit{bert-large-cased} \cite{devlin2019bert} and \textit{fBERT} \cite{sarkar2021fbert}. The \textit{bert-large-cased} is a general purpose pre-trained transformer model while \textit{fBERT} is a domain-specific pre-trained transformer model for offensive language identification that has been trained on over $1.4$ million offensive tweets in SOLID dataset \cite{rosenthal2020} and has shown state-of-the-art results in several offensive language identification benchmarks \cite{sarkar2021fbert}.

\vspace{2mm}
\noindent \textbf{\textit{Model Fusion}}
In order to combine the different models created using different datasets, we followed a recent approach named model fusion \cite{choshen2022fusing}. Model Fusion is the process of taking several fine-tuned models and creating a new base model. Formally, given an initialization base model $P$ and $n$ models fine-tuned on it, let $W_1,W_2\ldots W_n\in\mathbb{R}^d$ be the weights fine-tuned by the models over $P$. Fusing is a function

\resizebox{.85\columnwidth}{!}{
  \begin{minipage}{\columnwidth}
  \begin{align}
    W_{fuse} &= f(W_1, W_2, \ldots , W_n) 
    &\mathbb{R}^d \times \mathbb{R}^d \times \ldots \times \mathbb{R}^d  \rightarrow \mathbb{R}^d
\end{align}
  \end{minipage}
}

In this work, we propose the simplest form of fusion. For each weight shared by all models, assign the average weight to the model.

\resizebox{.85\columnwidth}{!}{
  \begin{minipage}{\columnwidth}
  \begin{align}
    W_{fuse} &=  f\left(W_1,W_2, \ldots , W_n\right) 
    &  =  \frac{W_1 + W_2 + \ldots + W_n}{n}
\end{align}
  \end{minipage}
}

In order to empirically evaluate model fusion in offensive language identification, we consider all possible seven combinations. These include different combinations of two models, such as $AHSD + OLID$ and $HASOC + HateX$, different combinations of three models, such as $AHSD + OLID + HASOC$ and $AHSD + OLID + HASOC$ and finally, the combination of all four models.

\vspace{2mm}
\noindent \textbf{\textit{Further Finetunning}}
The weights of the fused model resulting from step 2  can be anomalous as we followed a naive averaging method. Therefore, we performed a further finetuning step on the fused model. In this step, we fine-tuned the fused model using only one available dataset in a particular environment. We followed the same classification objective described in step 1 and used the same configurations. However, to avoid the model being biased toward the finetuning dataset, we only used $20\%$ of the available training data in the finetuning step.

\vspace{2mm}
\noindent The whole pipeline described above simulates a real-life scenario where the data can not be shared. The machine learning models are trained in separate environments using their own data, as in the first step. In the second step, with model fusion, we combined the models. In the final step. We further fine-tuned the fused model on a particular dataset where we repeated the process for all four datasets. Therefore with this pipeline, the datasets are not shared, and privacy is preserved among the different environments.

\subsection{Baseline Models}

We compared our fusion-based approach to two baseline models.  

\vspace{2mm}
\noindent \textbf{\textit{Non-fused Baseline}} We train a transformer-based baseline using the training set of one of the datasets and evaluate it on the test set of that particular dataset as well as on the test sets of other datasets. We repeated the process for all four datasets with two transformer models; \textit{bert-large-cased} \cite{devlin2019bert} and \textit{fBERT} \cite{sarkar2021fbert}. This baseline reflects the most common approach in offensive language detection, where a model is trained on a dataset available for a particular environment, but evaluated on other datasets in different environments as well.  

\noindent \textbf{\textit{Ensemble Baseline}} We also used an ensemble baseline; where we trained four separate transformer models on each dataset. For each test instance, we predicted values from all four models, and the final label is the label predicted with the highest probability from all four models. Similar to our previous experiments we repeated the process for \textit{bert-large-cased} \cite{devlin2019bert} and \textit{fBERT} \cite{sarkar2021fbert}. 


\section{Results and Discussion}

In Table \ref{tab:best_results}, we present the best results from each approach for each dataset. We show the results for fBERT as it provided better overall results. For the AHSD test set, the best result, $0.921$ Macro F1 score, is obtained when fBERT models are trained on AHSD and OLID and fused, then further fine-tuned on AHSD. For OLID the best result, $0.839$ Macro F1 score was provided when BERT-large-cased models trained on AHSD and OLID were fused and further fine-tuned on AHSD. Similarly, for HateX the best result, $0.777$ was provided when the fBERT models trained on AHSD and HateX were fused and further fine-tuned on HateX. However, HASOC follows a different pattern, and the best result was produced when fBERT models trained on AHSD, OLID and HASOC were fused, and further fine-tuned on AHSD. Overall, fBERT models provided slightly better results than BERT-large-cased models in most experiments. This is mainly because the fBERT model was trained on domain-specific data on offensive language identification. Finally, we present all results of the fused models and the non-fused model baseline in Table \ref{tab:results} in terms of Macro F1 score.

\subsection{Discussion}

\begin{table*}[ht]
    \centering
    \scalebox{0.62}{
        \begin{tabular}{c | c c c c | c c c c | c c c c }
        \hline
             \multirow{2}{*}{\shortstack{\bf Fine-tuned \\ \bf Dataset}} & & \multicolumn{2}{c}{\bf Fused Models} & & \multicolumn{4}{c}{\bf BERT-large-cased} & \multicolumn{4}{c}{\bf fBERT} \\
            \cmidrule(r){6-9}\cmidrule(r){10-13}
             & & & & & \bf AHSD & \bf OLID & \bf HASOC & \bf HATEX & \bf AHSD & \bf OLID & \bf HASOC & \bf HATEX \\
            \hline
            \multirow{8}{*}{AHSD} & AHSD & OLID & - & - & 0.900\textpm0.00 & 0.830\textpm0.07 & 0.610\textpm0.00 & 0.554\textpm0.06 & \bf 0.921\textpm0.00 & 0.836\textpm0.09 & 0.627\textpm0.00 & 0.628\textpm0.00\\
    
            & AHSD & - & HASOC & - & 0.778\textpm0.14 & 0.627\textpm0.00 & 0.637\textpm0.00 & 0.607\textpm0.02 & 0.776\textpm0.04 & 0.722\textpm0.00 & 0.632\textpm0.00 & 0.677\textpm0.05\\
            
            & AHSD & - & - & HATEX & 0.727\textpm0.03 & 0.697\textpm0.00 & 0.660\textpm0.04 & 0.594\textpm0.00 & 0.781\textpm0.03 & 0.707\textpm0.00 & 0.673\textpm0.03 & 0.648\textpm0.00\\
            \cmidrule(r){2-5}\cmidrule(r){6-9}\cmidrule(r){10-13}
            & AHSD & OLID & HASOC & - & \bf 0.919\textpm0.00 &  0.837\textpm0.08 & \bf 0.766\textpm0.02 & 0.636\textpm0.00 & 0.915\textpm0.00 & 0.835\textpm0.08 & \bf 0.770\textpm0.01 & 0.623\textpm0.00\\
            
            & AHSD & - & HASOC & HATEX & 0.705\textpm0.06 & 0.674\textpm0.00 & 0.595\textpm0.03 & 0.565\textpm0.00 & 0.734\textpm0.03 & 0.704\textpm0.00 & 0.643\textpm0.00 & 0.643\textpm0.00\\
            
            & AHSD & OLID & - & HATEX & 0.905\textpm0.00 & 0.813\textpm0.09 & 0.628\textpm0.00 & 0.719\textpm0.05 & 0.914\textpm0.00 & 0.834\textpm0.08 & 0.627\textpm0.00 & 0.772\textpm0.01\\
            \cmidrule(r){2-5}\cmidrule(r){6-9}\cmidrule(r){10-13}
            & AHSD & OLID & HASOC & HATEX & 0.716\textpm0.03 & 0.708\textpm0.00 & 0.646\textpm0.05 & 0.652\textpm0.06 & 0.730\textpm0.01 & 0.724\textpm0.00 & 0.668\textpm0.04 & 0.684\textpm0.04\\
             \cmidrule(r){2-5}\cmidrule(r){6-9}\cmidrule(r){10-13}
            & \multicolumn{3}{c}{\bf Non-fused Baseline} & & \underline {0.926\textpm0.01} & 0.699\textpm0.03 & 0.630\textpm0.05 & 0.586\textpm0.06 &  \underline {0.931\textpm0.01} & 0.743\textpm0.03 & 0.682\textpm0.04 & 0.606\textpm0.06 \\
            \midrule
            \multirow{8}{*}{OLID} & AHSD & OLID & - & - & 0.893\textpm0.00 & \bf 0.839\textpm0.05 & 0.647\textpm0.00 & 0.621\textpm0.03 & 0.866\textpm0.00 & \bf 0.837\textpm0.03 & 0.601\textpm0.00 & 0.598\textpm0.00\\
    
            & - & OLID & HASOC & - & 0.715\textpm0.00 & 0.405\textpm0.01 & 0.392\textpm0.00 & 0.651\textpm0.06 & 0.718\textpm0.00 & 0.725\textpm0.07 & 0.655\textpm0.00 & 0.667\textpm0.05\\
            
            & - & OLID & - & HATEX & 0.696\textpm0.00 & 0.692\textpm0.08 & 0.656\textpm0.04 & 0.616\textpm0.00 & 0.679\textpm0.07 & 0.723\textpm0.07 & 0.611\textpm0.00 & 0.650\textpm0.00\\
            \cmidrule(r){2-5}\cmidrule(r){6-9}\cmidrule(r){10-13}
            & AHSD & OLID & HASOC & - & 0.868\textpm0.00 & 0.826\textpm0.04 & 0.756\textpm0.00 & 0.608\textpm0.00 & 0.840\textpm0.00 & 0.819\textpm0.02 & 0.759\textpm0.09 & 0.606\textpm0.00\\
            
            & - & OLID & HASOC & HATEX & 0.687\textpm0.00 & 0.649\textpm0.09 & 0.586\textpm0.01 & 0.596\textpm0.00 & 0.729\textpm0.00 & 0.694\textpm0.08 & 0.637\textpm0.01 & 0.630\textpm0.00 \\
            
            & AHSD & OLID & - & HATEX & 0.847\textpm0.00 & 0.812\textpm0.04 & 0.642\textpm0.00 & 0.751\textpm0.09 & 0.861\textpm0.00 & 0.831\textpm0.03 & 0.615\textpm0.00 & 0.752\textpm0.01 \\
            \cmidrule(r){2-5}\cmidrule(r){6-9}\cmidrule(r){10-13}
            & AHSD & OLID & HASOC & HATEX & 0.713\textpm0.00 & 0.777\textpm0.00 & 0.672\textpm0.07 & 0.699\textpm0.08 & 0.708\textpm0.08 & 0.793\textpm0.00 & 0.682\textpm0.08 & 0.707\textpm0.09 \\
              \cmidrule(r){2-5}\cmidrule(r){6-9}\cmidrule(r){10-13}
            & \multicolumn{3}{c}{\bf Non-fused Baseline} & & 0.685\textpm0.02 & \underline {0.845\textpm0.00} & 0.636\textpm0.05 & 0.620\textpm0.06 & 0.702\textpm0.01 & \underline {0.851\textpm0.00} & 0.653\textpm0.05 & 0.645\textpm0.08 \\
            \midrule
            \multirow{8}{*}{HASOC} & AHSD & - & HASOC & - & 0.777\textpm0.13 & 0.419\textpm0.00 & 0.652\textpm0.00 & 0.356\textpm0.06 & 0.792\textpm0.11 & 0.785\textpm0.05 & 0.680\textpm0.00 & 0.708\textpm0.08\\
    
            & - & OLID & HASOC & - & 0.147\textpm0.00 & 0.707\textpm0.05 & 0.656\textpm0.00 & 0.220\textpm0.07 & 0.717\textpm0.00 & 0.734\textpm0.05 & 0.683\textpm0.00 & 0.673\textpm0.04\\
            
            & - & - & HASOC & HATEX & 0.530\textpm0.05 & 0.480\textpm0.00 & 0.695\textpm0.04 & 0.738\textpm0.00 & 0.761\textpm0.03 & 0.791\textpm0.00 & 0.689\textpm0.00 & 0.690\textpm0.00 \\
            \cmidrule(r){2-5}\cmidrule(r){6-9}\cmidrule(r){10-13}
            & AHSD & OLID & HASOC & - & 0.864\textpm0.00 & 0.812\textpm0.05 &  0.763\textpm0.08 & 0.624\textpm0.00 & 0.805\textpm0.00 & 0.801\textpm0.00 & 0.754\textpm0.07 & 0.635\textpm0.00\\
            
            & AHSD & - & HASOC & HATEX & 0.754\textpm0.01 & 0.419\textpm0.00 & 0.686\textpm0.01 & 0.698\textpm0.00 & 0.734\textpm0.09 & 0.780\textpm0.00 & 0.668\textpm0.01 & 0.661\textpm0.00 \\
            
            & - & OLID & HASOC & HATEX & 0.732\textpm0.00 & 0.700\textpm0.04 & 0.675\textpm0.01 & 0.686\textpm0.00 & 0.736\textpm0.00 & 0.712\textpm0.06 & 0.671\textpm0.00 & 0.676\textpm0.00 \\
            \cmidrule(r){2-5}\cmidrule(r){6-9}\cmidrule(r){10-13}
            & AHSD & OLID & HASOC & HATEX & 0.703\textpm0.09 & 0.647\textpm0.00 & 0.651\textpm0.00 & 0.651\textpm0.00 & 0.719\textpm0.06 & 0.781\textpm0.00 & 0.702\textpm0.06 & 0.718\textpm0.06\\
             \cmidrule(r){2-5}\cmidrule(r){6-9}\cmidrule(r){10-13}
            & \multicolumn{3}{c}{\bf Non-fused Baseline} & & 0.620\textpm0.03 & 0.492\textpm0.01 & \underline {0.788\textpm0.01} & 0.555\textpm0.06 &  0.645\textpm0.02 & 0.532\textpm0.01 & \underline {0.798\textpm0.01} & 0.575\textpm0.05 \\
            \midrule
            \multirow{8}{*}{HATEX} &  AHSD & - & - & HATEX & 0.758\textpm0.01 & 0.449\textpm0.00 & 0.531\textpm0.08 & 0.744\textpm0.00 & 0.671\textpm0.01 & 0.591\textpm0.00 & 0.587\textpm0.00 & \bf 0.777\textpm0.00\\
    
            &  - & OLID & - & HATEX & 0.650\textpm0.00 & 0.689\textpm0.06 & 0.557\textpm0.09 & 0.749\textpm0.00 & 0.584\textpm0.02 & 0.668\textpm0.01 & 0.599\textpm0.00 & 0.775\textpm0.00\\
            
            & - & - & HASOC & HATEX & 0.538\textpm0.01 & 0.545\textpm0.0 & 0.710\textpm0.05 & \bf 0.756\textpm0.00 & 0.527\textpm0.05 & 0.573\textpm0.00 & 0.707\textpm0.07 & 0.772\textpm0.00\\
            \cmidrule(r){2-5}\cmidrule(r){6-9}\cmidrule(r){10-13}
            & AHSD & - & HASOC & HATEX & 0.692\textpm0.04 & 0.529\textpm0.00 & 0.693\textpm0.05 & 0.741\textpm0.00 & 0.636\textpm0.10 & 0.588\textpm0.00 & 0.688\textpm0.08 & 0.767\textpm0.00\\
            
            & - & OLID & HASOC & HATEX & 0.561\textpm0.00 & 0.640\textpm0.09 & 0.690\textpm0.06 & 0.755\textpm0.00 & 0.526\textpm0.00 & 0.664\textpm0.08 & 0.689\textpm0.08 & 0.772\textpm0.00\\
            
            & AHSD & OLID & - & HATEX & 0.522\textpm0.00 & 0.597\textpm0.08 & 0.607\textpm0.00 & 0.645\textpm0.09 & 0.532\textpm0.00 & 0.563\textpm0.03 & 0.613\textpm0.00 & 0.633\textpm0.10\\
            \cmidrule(r){2-5}\cmidrule(r){6-9}\cmidrule(r){10-13}
            & AHSD & OLID & HASOC & HATEX & 0.627\textpm0.08 & 0.532\textpm0.00 & 0.635\textpm0.09 & 0.642\textpm0.11 & 0.631\textpm0.09 & 0.565\textpm0.00 & 0.652\textpm0.09 & 0.671\textpm0.11 \\
              \cmidrule(r){2-5}\cmidrule(r){6-9}\cmidrule(r){10-13}
            & \multicolumn{3}{c}{\bf Non-fused Baseline} & & 0.569\textpm0.03 & 0.504\textpm0.01 & 0.604\textpm0.02 & \underline {0.782\textpm0.02}  &  0.581\textpm0.01 & 0.523\textpm0.01 & 0.612\textpm0.01 & \underline {0.795\textpm0.01} \\
            \midrule
        \end{tabular}
    }
    \caption{Macro F1 score results for the fuse models (BERT-large-cased and fBERT) compared to the baseline systems fine-tuned on the four datasets. Results are reported on 10 runs along with standard deviation. The best results from the fused approach for each model are in bold. Results for the non-fused baseline model  evaluated on the same dataset are underlined.}
    \label{tab:results}
\end{table*}

\noindent We discuss the following four main findings from our results; 

\vspace{3mm}

\noindent \textbf{\textit{(1) The fused model performs better when evaluated on the same dataset used in further finetunning.}} All the datasets except for HASOC, the best result was produced when the fused model was further fine-tuned on that particular dataset. For HASOC too, when the fBERT model trained on AHSD, OLID and HASOC were fused and further fine-tuned on HASOC provided 0.754 Macro F1 score, which is very close to the best result (0.770). With the results, we can conclude that the fused model performs better when evaluated on the same dataset used in further finetunning.  This observation reflects an ideal scenario in real-world applications where we want an ML model to perform excellently in data specific to our environment/ platform. This objective can be achieved successfully with model fusion and finetunning as we see in the results. 

\vspace{3mm}

\noindent \textbf{\textit{(2) The fused model generalizes well across datasets even when it is not used in finetunning.}} One drawback of fused models is that the result slightly decreases compared to the non-fused models trained only using a particular dataset. In the results, this is clear as there is a decrease in the Macro F1 score between underlined values and bolded values. Furthermore as you can see in Table \ref{tab:best_results}, the best result in all the datasets were produced with the non-fused baseline. However, after further investigating this, it is clear that non-fused models do not often generalise well across other datasets. For example in Table \ref{tab:results}, the non-fused model trained on AHSD only provides 0.699 Macro F1 score for OLID. However, AHSD and OLID fused model further fine-tuned on AHSD provides 0.830 Macro F1 score. This is similar to the majority of the experiments, and fused models provide better results than non-fused models in other datasets. This observation again reflects an ideal scenario in real-world applications where we want an ML model to perform well across data not specific to our environment/ platform. As we see in the results, this objective can be achieved successfully with model fusion. 

\vspace{3mm}

\noindent \textbf{\textit{(3) The Fused model outperforms the ensemble baseline in all the datasets.}}
As shown in Table \ref{tab:best_results}, model fusion approaches with and without finetunning on a particular dataset outperform the best ensemble model. For HASOC, there is a large gap between the ensemble model and fused models as the ensemble model produces only 0.670 Macro F1 score while the fused model provides 0.770 Macro F1 score. The other datasets also follow a similar pattern. This is a key observation because we have presented a fusion based approach for FL that can surpass an ensemble based model preserving privacy across different datasets. The platforms/ environments that are interested in developing a FL approach should focus on model fusion based strategies that outperform ensemble based models as we showed in the results. 

\vspace{3mm}

\noindent \textbf{\textit{(4) The Fused model performance heavily depends on the datasets it was trained on.}}
Our final observation is that the fused model performance depends on the datasets that it was trained on. For example, when the model fusion was performed between AHSD and OLID, the final model provided excellent results on both datasets. This is due to the general nature of these two datasets covering multiple types of offensive content rather than focusing on a particular type of offensive content. On the other hand, results are not the same when the model fusion was performed between AHSD and HASOC where the final model did not provide good results for both datasets.  This can be explained by the demography of the dataset as HASOC data is collected on Twitter users based in India. It is clear that model fusion would thrive in similar kinds of datasets, but would not perform well with different kinds of data. 

\vspace{2mm}

\noindent Overall, model fusion produces excellent results on the dataset that it was fine-tuned on, and it generalizes well across other datasets. Fused models outperform both of our baselines in all the datasets. Therefore, model fusion provides a successful approach to FL.

\subsection{Multilingual Experiments}

We conducted initial multilingual experiments with the same FL setting. We used OffendES \cite{plaza2021offendes}, a Spanish offensive language identification dataset. For English we used the OLID dataset described before. Each instance in OffendES is labelled as belonging to one of the five classes; Offensive and targeted to a person (OFP), Offensive and targeted to a group (OFG), Offensive and not targeted to a person or a group (OFO), Non-offensive, but with expletive language (NOE), and Non-offensive (NO). We map the instances belonging to the OFP, OFG, OFO, and NOE to OLID OFF, and the NO class as NOT. Even though, the label NOE is considered non-offensive in OffendES, it contains profanity so we map it to OLID label OFF to conform with the OLID guidelines.

Instead of the monolingual BERT models we used in the previous experiments, we use cross-lingual models, specifically XLM-R \cite{conneau2019unsupervised}. We used the same FL settings and compared it with ensemble baseline. The results are shown in Table \ref{tab:multilingual}.

\begin{table}[ht!]
    \centering
    \scalebox{0.92}{
        \begin{tabular}{c | c | l  }
        \hline
             \bf Dataset & \bf Approach & \bf Macro F1 \\

            \hline
                 \multirow{4}{*}{English} & non-fused & 0.845 \textpm 0.01\\
            & fusion with FT  & 0.829 \textpm 0.03\\
            & fusion without FT  & 0.831 \textpm 0.00\\
            & ensemble &    0.776 \textpm 0.02\\

          \hline
             \multirow{4}{*}{Spanish} & non-fused & 0.812 \textpm 0.04\\
            & fusion with FT  & 0.809 \textpm 0.02\\
            & fusion without FT  & 0.792 \textpm 0.01\\
            & ensemble &    0.761 \textpm 0.02\\

          \hline

        \end{tabular}
    }
    \caption{The results for multilingual experiments on English and Spanish; non-fused models, fused models with fine-tuning (FT), fused models without finetuning and ensemble. We report the results with xlm-roberta. The results are ordered from Macro F1.}
    \label{tab:multilingual}
\end{table}

\noindent The results show that fusion based FL outperforms ensemble baseline in multilingual settings too. This opens new avenues for privacy preserving models for languages other than English and more specifically, low-resource languages.

\section{Conclusion and Future Work}

This paper introduced FL in the context of combining different offensive language identification models. While a recent study \cite{gala2023federated} uses FL learning in offensive language identification, their work is limited to distributed training on the same dataset with multiple clients. As far as we know, our research is the first study to use FL in combining multiple offensive language identification models. We evaluated a fusion-based FL architecture using a general BERT model and a fine-tuned fBERT model on four publicly available English benchmark datasets. We also presented initial cross-lingual experiments in English and Spanish. Our results show that the fusion model performances outperform the performance of an ensemble baseline model. We also show that the fused model generalizes well across all datasets tested. As the FL architecture does not require data sharing, we believe that FL is a promising research direction in offensive language identification due to its privacy preserving nature. 

In future work, we would like to explore other FL architectures and compare their performance to the fused model proposed in this paper. Finally. we would like to evaluate the performance of recently proposed large language models (LLMs) (e.g., GPT-4, LLama 2) for this task in FL settings. 

\section*{Bibliographical References}\label{sec:reference}
\bibliographystyle{lrec-coling2024-natbib}
\bibliography{lrec-coling2024-example}

\end{document}